\documentclass[letterpaper, 10 pt, conference]{ieeeconf}  

\IEEEoverridecommandlockouts                              
\usepackage{amssymb}
\usepackage{amsmath}
\usepackage{graphics}
\usepackage{epsfig}
\usepackage{multirow}
\usepackage{nicefrac}
\usepackage{cases}          
\usepackage{subcaption}
\usepackage{mathtools}
\usepackage{xcolor}
\usepackage{algorithm}
\usepackage{algorithmicx}
\usepackage{algpseudocode}
\usepackage{balance}

\newcommand{\tn}[1]{\textnormal{#1}}
\newcommand{\tb}[1]{\textbf{#1}}

\newcommand{\mat}[0]{\begin{bmatrix}}
\newcommand{\mate}[0]{\end{bmatrix}}

\newcommand{\vf}{\mathbf{f}}
\newcommand{\vg}{\mathbf{g}}

\newcommand{\vp}{\mathbf{p}}

\newcommand{\vu}{\mathbf{u}}
\newcommand{\vv}{\mathbf{v}}

\newcommand{\vx}{\mathbf{x}}

\newcommand{\cA}{\mathcal{A}}

\newcommand{\cI}{\mathcal{I}}

\newcommand{\cO}{\mathcal{O}}

\newcommand{\cT}{\mathcal{T}}
\newcommand{\cU}{\mathcal{U}}

\newcommand{\cW}{\mathcal{W}}
\newcommand{\cX}{\mathcal{X}}

\newcommand{\cZ}{\mathcal{Z}}
\newcommand{\R}{\mathbb{R}}
\newcommand{\N}{\mathbb{N}}

\newcommand\norm[1]{\left\|#1\right\|}              

\newcommand{\vect}[1]{\boldsymbol{#1}}

\title{\LARGE \bf 
With Whom to Communicate: Learning Efficient Communication \\ for Multi-Robot Collision Avoidance
}

\author{\'{A}lvaro Serra-G\'{o}mez$^{1}$, Bruno Brito$^{1}$, Hai Zhu$^{1}$, Jen Jen Chung$^{2}$, Javier Alonso-Mora$^{1}$ 
\thanks{$^1$Department of Cognitive Robotics, Delft University of Technology}
\thanks{{\tt\small $\{$a.serragomez; bruno.debrito; h.zhu; j.alonsomora$\}$@tudelft.nl}}
\thanks{$^2$Autonomous Systems Lab, ETH Zurich}
\thanks{{\tt\small $\{$jenjen.chung$\}$ @mavt.ethz.ch}}
\thanks{This work is supported by the U.S. Office of Naval Research Global(ONRG) NICOP-grant N62909-19-1-2027}
}

\usepackage{graphicx}

\begin{document}

\maketitle
\thispagestyle{empty}
\pagestyle{empty}

\begin{abstract}
Decentralized multi-robot systems typically perform coordinated motion planning by constantly broadcasting their intentions as a means to cope with the lack of a central system coordinating the efforts of all robots. Especially in complex dynamic environments, the coordination boost allowed by communication is critical to avoid collisions between cooperating robots. However, the risk of collision between a pair of robots fluctuates through their motion and communication is not always needed. Additionally, constant communication makes much of the still valuable information shared in previous time steps redundant. This paper presents an efficient communication method that solves the problem of ``when" and with ``whom" to communicate in multi-robot collision avoidance scenarios. In this approach, every robot learns to reason about other robots' states and considers the risk of future collisions before asking for the trajectory plans of other robots. We evaluate and verify the proposed communication strategy in simulation with four quadrotors and compare it with three baseline strategies: non-communicating, broadcasting and a distance-based method broadcasting information with quadrotors within a predefined distance.

\end{abstract}

\section{Introduction}

Being able to account for the planned path of other robots is of utmost importance for safe navigation in Micro Aerial Vehicle (MAV) environments. Centralized systems achieve this objective by having a central robot manage all of the robots' information and plans. Instead, in decentralized systems robots estimate or communicate their teammates' future trajectories. Common communication policies are broadcasting or distance-based communication of trajectory plans. However, much of this information becomes redundant or unnecessary when robot motions are clearly not intersecting. This is inefficient and sometimes unfeasible, especially in communication-restrictive environments such as underwater, outer space or for large groups of robots. In this work we focus on the following two issues: a) solving the problem of \textit{when} and with \textit{whom} to communicate and b) how to couple this communication policy with existing motion planning methods.

In this paper we propose an efficient communication policy method combined with an optimal control motion planner for multi-robot collision avoidance. The approach leverages the strengths of learning methods for decision-making and nonlinear receding horizon control, or Non-Linear Model Predictive Control (NMPC) for multi-robot path planning. In particular, we use Multi-Agent Reinforcement Learning (MARL) to learn the robots' communication policies. For every robot and time instance, the policy selects a set of other robots and requests their trajectory plans. Non-selected robots are assumed to follow a constant velocity trajectory or their previously communicated one. Then, we formulate a nonlinear optimization problem to generate a safe trajectory. The planned trajectory takes into account the requested and estimated trajectories represented as constraints in the receding horizon framework.

The main contributions of this work are:
\begin{itemize}
    \item A combined communication policy and trajectory planning method for micro-aerial vehicles (MAVs), which utilizes the strengths of non-linear model predictive control (NMPC) and multi-agent reinforcement learning (MARL) to plan safe trajectories with minimal communications in three-dimensional scenarios.
    \item An on-line efficient communication policy that uses Multi-Agent Reinforcement Learning (MARL) to learn (off-line) \textit{when} and with \textit{whom} it is useful to communicate, performing collision avoidance while minimizing communication.
\end{itemize}

We evaluate our method with a team of quadrotors in simulated scenarios requiring different levels of communication for safe navigation and compare it to three other heuristic baseline methods. We show that our learning method enables the emergence of intuitive communication behaviours while maintaining the performance of broadcasting policies.

\section{Related Work}
\subsection{Communication in Collision Avoidance}
There has been a large amount of work in multi-robot collision avoidance. One of the approaches is the reciprocal velocity obstacles (RVO) method~\cite{VandenBerg2008}. From the basic RVO framework there are now several extensions: the optimal reciprocal collision-avoidance (ORCA) method~\cite{VanDenBerg2011} casts the problem into a linear programming formulation; the generalized RVO method~\cite{Bareiss2015} applies to heterogeneous teams of robots; and the $\varepsilon$-cooperative collision avoidance ($\varepsilon$CCA) method~\cite{Alonso-Mora2018} accounts for the cooperation of nonholonomic robots. While these RVO-based methods are computationally efficient, the robot dynamics are not fully modeled and the robot motion is typically limited by only planning one time step ahead. These issues can be overcome by using a model predictive control (MPC) framework for collision-free trajectory planning. This includes the decentralized MPC~\cite{Kamel2017, Arul2020RAL} assuming other robots are moving at a constant velocity, the distributed MPC~\cite{Luis2020RAL} with on-demand collision avoidance, and the chance-constrained MPC~\cite{Zhu2019RAL} that accounts for robot localization and sensing uncertainties. In this paper, we study the multi-robot collision avoidance problem in the MPC-based framework.

Typically in multi-robot collision avoidance, robots are assumed to be able to observe other robots' positions and estimate their velocity using a filter. However, each robot's intentions and planned trajectories are not known by other robots. One approach to tackle this issue is to let each robot communicate its planned trajectory with every other robot in the team. Robot's can then update their own trajectories to be collision free with other robots' trajectory plans, e.g. as in these distributed MPC works~\cite{Luis2020RAL, Zhu2019RAL, Ferranti2018ECC}. While these methods can achieve efficient and safe collision avoidance, the communication burden across the team is huge, particularly when the number of robots is large and much of the communication between robots may be redundant and unnecessary. Without communicating trajectory plans, a robot can achieve collision avoidance by constraining its motion to be within a safe neighborhood computed using only other robots' current positions, e.g. the BVC method~\cite{Zhou2017} and the B-UAVC method~\cite{Zhu2019MRS}. While these methods can guarantee collision avoidance without inter-robot communication, the planned robot motions are very conservative and inefficient. Alternatively, the decentralized MPC in~\cite{Kamel2017, Arul2020RAL} employs a constant velocity model when predicting other robots' future trajectories. While communication among robots are not required, the planned robot motions are not safe, in particular when the robots are moving at a high speed~\cite{Zhu2019RAL}.

\subsection{Communication Scheduling}

A lot of works tend to formulate the problem of efficient communication in a receding horizon fashion. Some methods formulate the problem as a decentralized version of a Markov Decision Process (Dec-MDP)~\cite{Roth2005} or Partially Observable MDP (Dec-POMDP)~\cite{Becker2009} and try to optimize a value function in which communications are penalized. Others, such as~\cite{Kassir2016}, choose to formulate a constrained optimization problem where communications must be directly minimized while still guaranteeing data flow throughout the network. These approaches need us to be able to directly quantify a priori the value of communication, which is what we are trying to avoid. Recent work~\cite{Best2018} manages to avoid this by triggering communication whenever uncertainty over another agent's actions exceeds a threshold. Ultimately, however, receding horizon methods are limited by their prediction horizon and the need for evaluation heuristics, which can unintentionally bias the resulting communication processes. On the other hand, reinforcement learning methods may discover more general policies without the need for delicate hand-tuning.

\subsection{Learning Methods for Coordination}

One of the main issues of Multi-Agent Reinforcement Learning (MARL) is instability of the learning process caused by the non-stationarity that results from having different interacting policies learning at the same time. In order to deal with this problem successfully, recent work on MARL~\cite{Lowe2017} performs centralized training and decentralized execution. Such a method has been successfully applied in the field of non-communicating collision avoidance tasks~\cite{Everett2018a, Everett2019a}. Regarding tasks that require communication, several works have been published recently. Many of them focus on learning what content should be shared among agents, be it in the form of a composition of binary signals~\cite{Foerster2016d} and predefined symbols~\cite{Mordatch2018b}, policy hidden layers~\cite{Sukhbaatar2016e}, or by directly sharing parameters among agents~\cite{Gupta2017}. The most relevant to our work additionally focus on learning policies that are able to appropriately choose \textit{when} and with \textit{whom} to communicate. Jiang and Lu~\cite{Jiang2018} assign roles to every agent, making some of them in charge of organizing a common communication channel with their neighbours. However, regions where there is no agent with such a role are left without coordination capabilities. Instead, Das et al.~\cite{Das2019} present an end-to-end MARL algorithm that creates an attention module which chooses whom to establish bilateral communications with. Similarly, the method we present in this paper can also be considered as an attention module targeting other agents. However, we set our communications to be unilateral to promote asymmetrical behaviour. Additionally, we decouple the problem of communication and motion planning, allowing the combination of our method with existing and well-tested solutions for motion planning in collision avoidance tasks.

\section{Preliminaries}

In this paper, we address the problem of deciding when and with whom to communicate during a multi-robot collision avoidance task. Though the proposed formulation is intended to be general, we are inspired by the results obtained in~\cite{Zhu2019RAL}, which show how in a collision-avoidance scenario, methods that incorporate communication have a clear advantage over those that do not. We approach the information-sharing process as a MARL problem where the robots must learn to request information effectively. In this section, we set the context for our targeted communication process by providing an overview of the Non-Linear Model Predictive Control method used for motion control, as well as our MARL framework, introducing relevant notation for this work.

\subsection{Multi-Robot Collision Avoidance }
Consider a team of $n$ robots moving in a shared workspace $\cW \subseteq \R^3$, where each robot $i \in \cI = \{ 1,2,\dots, n \} \subset \N $ is modeled as an enclosing sphere with radius $r$. The dynamics of each robot $i \in \cI$ are described by a discrete-time equation as follows,
\begin{equation}\label{eq:nonDyn}
    \vx_i^{k+1} = \vf(\vx_i^k, \vu_i^k), \quad \vx_i^0 = \vx_i(0),
\end{equation}
where $\vx_i^k \in \cX \subset \R^{n_x}$ denotes the state of the robot, typically including its position $\vp_i^k$ and velocity $\vv_i^k$, and $\vu_i^k \in \cU \subset \R^{n_u}$ the control inputs at time $k$. $\cX$ and $\cU$ are the admissible state space and control space respectively. $\vx_i(0)$ is the initial state of robot $i$. Any pair of robots $i$ and $j$ from the group are mutually collision-free if $\norm{\vp_i^k - \vp_j^k} \geq 2r, \forall i\neq j \in \cI, \forall k = 0, 1, \dots $. Each robot has a given goal location $\vg_i$, which generally comes from some high-level path planner or is specified by some user.

The objective of multi-robot collision avoidance is to compute a local motion $\vu_i^k$ for each robot in the group, that respects its dynamics constraints, makes progress towards its goal location $\vg_i$ and is collision-free with other robots in the team for a short time horizon.

\subsection{Distributed Model Predictive Control }
The key idea of using distributed model predictive control to solve the multi-robot collision avoidance problem is to formulate it as a receding horizon constrained optimization problem. For each robot $i \in \cI$, a discrete-time constrained optimization formulation with $N$ time steps and planning horizon $\tau = N\Delta t$, where $\Delta t$ is the sampling time, is derived as follows,
\begin{equation}\label{eq:dmpc}
    \begin{aligned}
    \min\limits_{\vx_i^{1:N}, \vu_i^{0:N-1}}  ~~        
                            & \sum_{k=0}^{N-1} J_i^k(\vx_i^k, \vu_i^k) + J_i^N(\vx_i^N, \vg_i) \\
    \text{s.t.}    ~~            & \vx_i^0 = \vx_i(0), \\
                            & \vx_i^{k} = \vf(\vx_i^{k-1}, \vu_i^{k-1}), \\
                            & \norm{\vp_i^k - \vp_j^k} \geq 2r,\\ 
                            & \vu_i^{k-1} \in \cU, \quad
                            \vx_i^k \in \cX,\\
                            &\forall j \neq i\in\cI; \, \forall k\in \{1,\dots,N\}.
    \end{aligned}
\end{equation}
At each time step, each robot in the team solves online the constrained optimization problem \eqref{eq:dmpc} and then executes the first step control inputs, in a receding-horizon fashion. 

\subsection{With Whom to Communicate}
Note that for each robot to solve problem \eqref{eq:dmpc}, it has to know the future trajectory of other robots in the team. At time $t$, denote by ${\hat{\cT}}_{i,j}^t = \{ \vp_j^{t+1:t+N} \}|_t$ the trajectory of robot $j\in\cI, j\neq i$ that robot $i$ assumes and uses in solving the problem \eqref{eq:dmpc}, where the hat $\hat{}$ indicates that it is what robot $i$ knows about the other agent's trajectory. Further denote by $\cT_i^t = \{ \vp_i^{t+1:t+N} \}|_t$ the trajectory for robot $i$ planned at time $t$. Typically, there are two methods for robot $i$ to obtain the future trajectory information of other robots $j$:
\begin{itemize}
    \item \tb{Without communication:} robot $i$ predicts another robot's future trajectory based on their current states, that is
    \begin{equation}
        {\hat{\cT}}_{i,j}^t = \tn{prediction}(\vx_j^t),~~ \forall j\neq i \in \cI.
    \end{equation}
    In this paper, each robot employs a constant velocity model for the prediction as described in \cite{Kamel2017}.
    \item \tb{Full communication:} All robots in the team communicate their planned trajectories to each other at each time step, that is 
    \begin{equation}
        {\hat{\cT}}_{i,j}^t = \cT_j^{t-\Delta t}, ~~\forall j\neq i \in \cI.
    \end{equation}
\end{itemize}

Both of the two methods have their advantages and disadvantages. While the full communication method can achieve safe collision avoidance, it requires a large amount of communication among robots. However, if there is no communication, the robot may plan an unsafe trajectory if its prediction on other robots' trajectories deviates from their real ones. 

Motivated by these facts, this paper aims to solve the problem of ``with whom to communicate'' for each robot in the team for collision avoidance. More precisely, at each time step, each robot $i$ decides whether or not to request a communication message from every other robot $j$. If robot $i$ decides to request robot $j$, robot $j$ should communicate its planned trajectory to robot $i$. If robot $i$ decides not to request robot $j$, it predicts robot $j$'s future trajectory based on its observed current state of robot $j$. 

Denote by $\pi_i^{t} = \{ c_{i,j}^t|j = 1, \dots, n\}$ the communication vector of robot $i$ at time $t$, in which $c_{i,j}^t = 1$ indicates that robot $i$ requires a communicated trajectory from robot $j$. Otherwise $c_{i,j}^t = 0$. Note that $c_{i,i}^t = 0$ since the robot does not need to communicate with itself. Let $\pi^t = \{ \pi_1^t; \dots; \pi_n^t \}$ be the communication matrix of the multi-robot system at time $t$. We define the communication cost of the system to be
\begin{equation}
    C(\pi^t) = \sum_i^n\sum_{j}^n c_{i,j}^t.
\end{equation}
The objective of this paper is to find a policy for each robot $i$, 
\begin{equation}
    \pi_i^t = \pi_i(\vx_1^t, \vx_2^t, \dots, \vx_n^t),
\end{equation}
that minimizes $C(\pi^t)$ while ensuring that the robots are collision-free with each other in the system.

\section{Method}

\subsection{Overview}

\begin{figure}[h]
    \centering
    \includegraphics[width=8.5cm]{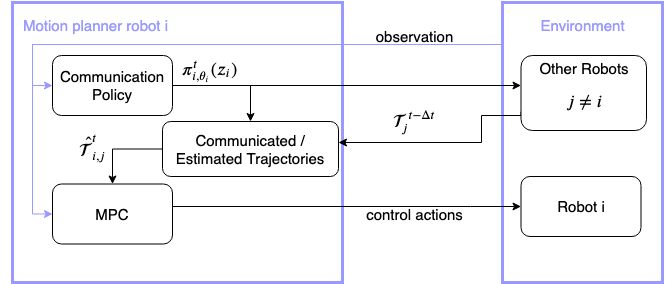}
    \caption{Schema of the proposed method for efficient communication. $\pi^t_i(z_i)$ is the communication policy dependent on the observation $z_i$. $\cT_j^{t-\Delta t}$ is the trajectory plan of robot j at the previous time step. And ${\hat{\cT}}_{i,j}^t$ are the combination of obtained and estimated trajectories of the other robots.}
    \label{fig:schema}
\end{figure}

An overview of the proposed method is given in Fig.\ref{fig:schema}. It consists of two components: a communication policy and a constrained MPC planner.

Every time step, based on its partial observation of the current joint state $z^t_i$, every robot targets a set of other robots $\pi^t_i$ and requests their trajectory plans ${\hat{\cT}}_{i,j}^t = \cT_j^{t-\Delta t}$ according to a learnt parametric policy $\pi_{i,\theta_i}(z^t_i)$. Those robots not targeted are estimated to follow a previously communicated trajectory or, in case it is no longer useful, a constant velocity model ${\hat{\cT}}_{i,j}^t = \tn{prediction}(\vx_j^t)$ as described in Section~III.C.

A receding horizon optimization is then employed to plan the future trajectory $\cT_i^t$ for robot $i$. To guarantee the safety of such a trajectory, the resulting trajectory is constrained to not intersect with the requested and estimated trajectories. The first action input from the computed plan is applied.


\subsection{Multi-robot Reinforcement Learning}

We formulate a multi-robot reinforcement learning problem to compute an efficient communication policy. By considering the optimization process as part of the transition model, this problem can be transformed into a decentralized POMDP \cite{Berstein2002}. The decentralized POMDP is composed of six components, including state space, action space, observation space, transition model, observation model.

\subsubsection{State Space $\cX$: }
For every robot $i$, $x^i \in \cX$ must account for the current physical state, its sequence of intended actions from the previous time step, as well as any information it has of other robots on their positions, velocities and action sequence intentions. Therefore, the state at time $t$ can be defined as:
\begin{equation}
    \vect{x}^t_i := [ \vect{p}^t_i, \vect{v}^t_i, \{ \vect{u}_i^{k'} \}_{k'= t-1,..,t+K-1}, \pi_i^{t-1} ],
\end{equation}
\begin{equation}
    \vect{X}^t := \{ x^t_1, x^t_2,..., x^t_n \},
\end{equation}
where $\vect{p}_i^t \in \mathbb{R}^3$ and $\vect{v}_i^t \in \mathbb{R}^3$ are the position and velocities of robot $i$ at time $t$, and $\{ u_i^{k'} \}_{k'= t-1,..,t+K-1}$ is the action sequence planned for a $K$-time-window at the previous time step $t-1$ by robot $i$. $\pi_i^{t-1} \in \{0,1\}^{n-1}$ is the binary vector indicating whether robot $i$ has requested any other robot $j\neq i$ trajectory intentions at time step $t-1$. Then, $\vect{X}^t$ is the joint state of the whole multi-robot system.


\subsubsection{Observation Space $\cZ$}

We assume each robot can always observe the positions and velocities of all other robots and knows the position of its goal through its sensors. For robot $i$, partial observations on the joint state at time $t$ are defined as follows:
\begin{equation}
    \vect{z}^t_i = [\vect{v}_i^t, \vect{p}_{i,g}^t, \{\vect{p}_{i,j}^t\}_{j\in I\backslash i}, \{\vect{v}_{i,j}^t\}_{j\in I\backslash i} ],
\end{equation}
where $\vect{p}_{i,j}^t$ and $\vect{v}_{i,j}^t$ are the relative positions and velocities of the other robots with respect to the $i^{th}$ robot, and $\vect{p}_{i,g}^t$ is the relative position of robot $i$'s goal. The joint observation from all robots is denoted by $\vect{z} = \{\vect{z}^1,...,\vect{z}^n\} \in \cZ$

\subsubsection{Action Space $\cA = \times_{i\in \cI}\cA_i$}
As it has already been introduced in Section~III.C, we denote by $\pi_i^{t} = \{ c_{i,j}^t|j \neq i \}$ the communication vector of robot $i$ at time $t$. Note we have dropped the $i^{th}$ element as the robot cannot communicate with itself. Therefore the action space for robot $i$ is:
$$\cA_i = \{0,1\}^{n-1}$$

\subsubsection{Reward $R(\vect{x}^t,\pi^t)$}
The reward function is chosen based on the behaviors we want to learn. It aims for the learned communication policy to communicate as little as possible while allowing each robot on the team to reach its goal and avoid collisions. The reward value $R(\vect{x}^t,\pi^t)$ is the immediate reward that all robots get at a state $x \in \cX$ after applying the communication matrix $\pi^t$. All robots getting the same reward accounts for indirect interactions e.g. robot $i$ colliding with another robot $j'$ (whose trajectory was not requested) because of trying to avoid the trajectory plan of robot $j$. The reward function is composed of the following weighted combination of terms:
\begin{equation}
    R(\vect{x}^t,\pi^t) = w_gR_g(\vect{x}^t) + w_{coll}R_{coll}(\vect{x}^t) + w_{c}R_{c}(\pi^t)
\end{equation}
where
\[   
R_g(\vect{x}^t) = 
     \begin{cases}
       r_g &  \forall i \in \cI, \norm{\vect{p}^t_{i,g}-\vect{p}^t_i} \leq r_i\\
       0 & \text{otherwise} \\
     \end{cases}
\]
where $r_g > 0$ is a tuned reward given for every time step that all robots are within its goal, $r_i$ is the radius of the smallest sphere containing the robot. The sooner all robots reach their destination, the more reward they receive during the episode, not only encouraging collision avoidance but also to reach the goal quickly.
\[   
R_{coll}(\vect{x}^t) = 
     \begin{cases}
       -r_{coll} &  \forall i,j\in \cI, i\neq j, \norm{\vect{p}^t_{i}-\vect{p}^t_j} \leq r_i+r_j\\
       0 & \text{otherwise} \\
     \end{cases}
\]
where $r_{coll} > 0$ is a tuned penalty term for the collision between any two robots.

Finally the global penalization term for path plan requests has been introduced before in Section~III.C and has the form:
$$R_{c}(\pi^t) = -C(\pi^t) = -\sum_i^n\sum_j^n c_{i,j}^t.$$

\subsubsection{Observation Model $\cO(\vect{z}^{t+1},\vect{x}^{t+1},\pi^t)$} We assume that every robot $i$ can directly observe the positions and velocities of other robots. The main uncertainty lies in their trajectory plans and their communication matrix $\pi^t_j$ .

\subsubsection{Transition model $T(\vect{x}^{t+1},\pi^t,\vect{x}^{t})$}
The transition model can be decomposed into a communication step and a physical action step:
\begin{equation}
    p(\vect{x}^{t+1}|\pi^t,\vect{x}^t) = p(\vect{x}^{t+1}|\vect{u}^t,\vect{x}^t)p(\vect{u}^t|\pi^t,\vect{x}^t)
\end{equation}
where $\vect{u}^t$ are the control actions applied at time step $t$, which are obtained from the motion planner. $p(\vect{u}^t|\pi^t,\vect{x}^t)$ models the effects of communication $\pi^t$ on the constrained optimization problem used to compute actions $\vect{u}^t$. Then, $p(\vect{x}^{t+1}|\vect{u}^t,\vect{x}^t)$ is the state transition for every robot. The robots employed in this paper are quadrotors, thus the state transition can be interpreted as the quadrotor model explained in Sec.III.A. 


\subsection{Multi-Agent Actor-Critic}
In order to find a policy $\pi_{\theta}$ maximizing a cost function $J(\theta) = \mathbb{E}_{\vect{x}\sim p_{\pi_{\theta}},a\sim \pi_{\theta}(\vect{x})}[R(\vect{x},a)]$ where $a\in \cA$, Policy Gradient methods (PG)~\cite{Sutton2000} directly adjust the parameters $\theta$ of the policy by taking steps in the direction of the gradient of $J$ with respect to the policy parameters:
\begin{equation}
\nabla_{\theta}J(\theta) = \mathbb{E}_{\vect{x}\sim p_{\pi_{\theta}},a\sim \pi_{\theta}(\vect{x})}[\nabla_{\theta}log\pi_{\theta}(a|\vect{x})Q_{\pi}(\vect{x},a)] 
\end{equation}
where $Q_\pi(x',a') = \mathbb{E}_{x\sim p_{\pi_{\theta}},a\sim \pi_{\theta}(x)}[R(x,a) | x',a']$ is the expected value from the total expected return conditioned on taking action $a'$ at state $x'$ and follow policy $\pi$ from then onwards. Actor-Critic methods are a family of algorithms that learn an approximation of the Q-function using deep neural networks. In particular, Deep Deterministic Policy Gradient (DDPG) algorithms~\cite{Lillicrap2014} are a variant of off-policy Actor-Critic methods that learn deterministic policies $\pi_{\theta}(a|x) = \mu_{\theta}(x)$ instead of stochastic ones and use another policy $\beta$ to explore the state-action space. Due to this modification, the expression of the gradient changes to the following:
\begin{multline}
  \nabla_{\theta^{\mu}}J = \mathbb{E}_{\vect{x}^t\sim\rho^{\beta}}[\nabla_aQ(\vect{x},a|\theta^Q)|_{\vect{x}=\vect{x}^t,a=\mu(\vect{x}^t)}\\
  \nabla_{\theta^{\mu}}\mu(\vect{x}|\theta^{\mu})|_{\vect{x}=\vect{x}^t}].
\end{multline}

The algorithm we are using in our approach is the natural extension of DDPG to multi-agent environments, that is, the Multi-Agent Deep Deterministic Policy Gradient algorithm (MADDPG)~\cite{Lowe2017}. While very similar to DDPG, MADDPG proposes for every agent to learn a decentralized policy with partial observations, while using its own centralized action-value function for learning using global state information, once again changing the expression for the gradient:
\begin{multline}
\nabla_{\theta_i}J(\mu_i) = \mathbb{E}_{\vect{x}^t\sim\rho^{\mu}}[\nabla_{a_i}Q_i^{\mu}(\vect{z},a_1,a_2,...,a_n)|_{a_i=\mu_i(\vect{z}_i)}\\\nabla_{\theta_i}\mu(\vect{z}_i|\theta_i)],
\end{multline}
where $\mathbf{z}$ consists of the observations of all agents $\vect{z} = (\vect{z}_1,...,\vect{z}_N)$. This technique allows us to cope with the non-stationarity resulting from having agents simultaneously learning interacting policies.

For every robot $i$, we want to learn a decentralized communication policy that targets other robots whose path plan is useful based on current observations. Thus, the policy of robot $i$ at time step $t$ will follow the expression:
\begin{equation}
    \pi_{i,\theta_i}(\vect{z}^t_i) = \{\mathbf{1}[f_{\theta_i}^{i,j}(\vect{z}^t_i) > \delta]\}_{j\in I\backslash i} = \pi_i^t \in \{0,1\}^{n-1},
\end{equation}
where $\pi_{i,\theta_i}(z_i)$ is the parameterized communication policy, $f_{\theta_i}^{i,j}$ is the parameterized function (e.g. neural network) mapping partial observations from robot $i$ to a communication score between $[0,1]$, $\delta$ is the threshold discriminating whether robot $i$ should request robot $j$'s path intentions and $\mathbf{1}$ is the indicator function. The value of $\delta$ can be chosen as a hyperparameter. To learn the communication policy using this method, careful consideration must be given to the exploration of the state-action space. This policy does not directly influence the state transitions, which makes the task of finding a good combination of communication requests through the episode complex. While it might pose a risk to use them in tasks where safety is a requirement, stochastic policies allow good exploration at training time making the task of finding efficient communication behaviors easier. 

The method used to learn this policy, MADDPG~\cite{Lowe2017}, is off-policy which means we can learn a deterministic target policy while using a stochastic policy at training time to encourage exploration.
To do so, we substitute the preset threshold by a $\delta'\sim U(0,1)$, which we sample at every time step. This enables us to apply exploration coherently as all scores attributed to other robots by robot $i$ during the same time step will be subject to the same sampled threshold. Requests from robot $i$ to robot $j$ following the exploration policy $\beta$ are denoted by:
\begin{equation}
    \beta_{i,\theta_i}(c^t_{i,j} | z^t_i) = \mathbb{P}_{\theta_i}( c^t_{i,j} = 1 | \vect{z}^t_i) = \mathbb{P}(f_{\theta_i}^{i,j}(\vect{z}_i^t) > \delta').
\end{equation}

\section{Results}
In this section we describe our implementation of the proposed method and evaluate it in simulation.  
\begin{figure*}[!t]
        \captionsetup[subfigure]{position=b}
        \begin{subfigure}{0.34\textwidth}
                \includegraphics[width=\textwidth]{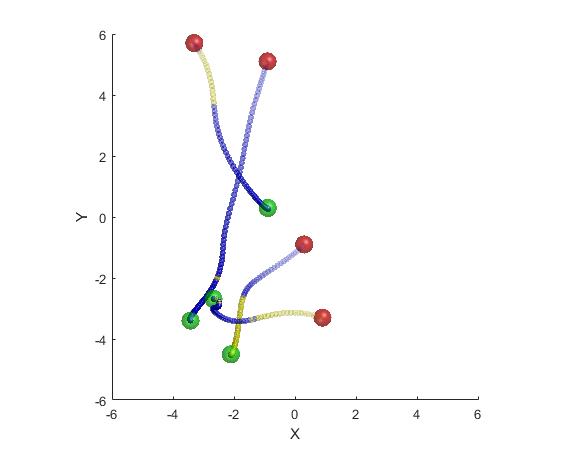}
        \caption{Random \label{subfig:scenario2}}
        \end{subfigure}
        ~
        \captionsetup[subfigure]{position=b}
        \begin{subfigure}{0.34\textwidth}
                \includegraphics[width=\textwidth]{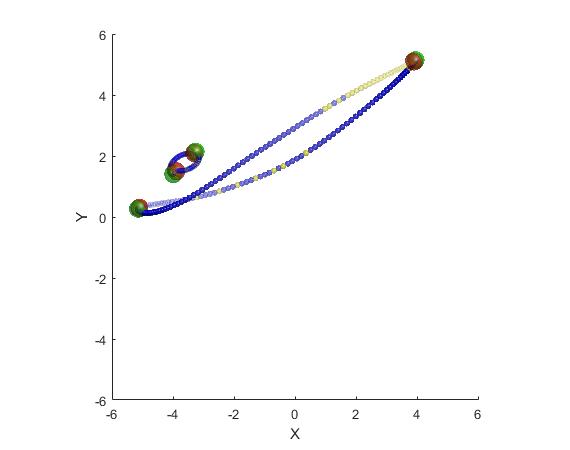}
        \caption{Random Swapping \label{subfig:scenario3}}
        \end{subfigure}
        ~
        \captionsetup[subfigure]{position=b}
        \begin{subfigure}{0.34\textwidth}
                \includegraphics[width=\textwidth]{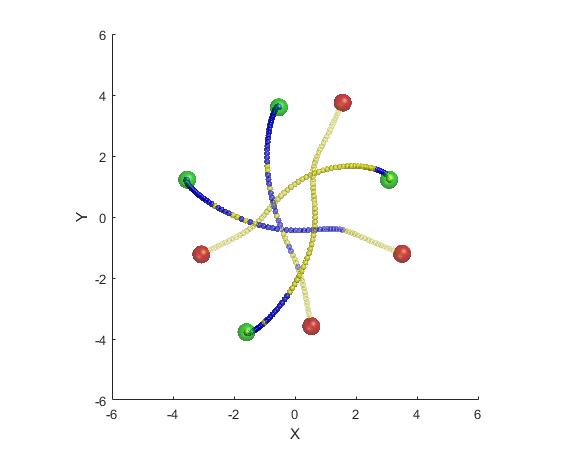}
        \caption{Asymmetric swapping \label{subfig:scenario4}}
        \end{subfigure}
        \caption{Simulation results for each scenario using our communication policy. Solid lines represent the trajectories executed by the drone-swarm. Yellow represents the positions where the drones communicate their trajectory plans. Blue depicts the positions where the drones do not communicate. Green and Red represent the initial and goal position of each drone, respectively. Increasing opacity represent the episode progression.}%
        \label{fig:scenarios}%
\end{figure*}
\subsection{Simulation Setup}
The simulation environment and NMPC controller were implemented in Matlab. We rely on the solver Forces Pro~\cite{domahidi2014forces} to generate optimized NMPC code. The learning algorithm was implemented in Python and ROS as middle-ware to connect both simulator and learning method. The Critic and Actor models are parameterized by two fully connected layers with 64 units and ReLu activation and were trained for 10000 episodes in an Intel i7 CPU@2.6GHz computer. We use the same hyperparameters reported in \cite{Lowe2017} for training except for $\gamma = 0.98$
Values for the reward weights were $w_g = 1$, $w_{coll} = 1$, $w_c = 0.1$. Tuned reward and penalty terms were $r_g = 1.3$, $r_{coll} = 150$. Episodes finished after reaching a collision or 100 time steps.
\subsection{Training Environment}\label{sec:scenarios}
We have created a simulation environment where a group of four drones navigate from an initial position to a goal position and must communicate their trajectory plans to perform collision avoidance. We have designed four different scenarios to evaluate our communication policy, as depicted in Fig.~\ref{fig:scenarios}. Each scenario has a different level of difficulty for the drone swarm to perform collision avoidance, ranging from a simple scenario where no communication is needed (e.g., Fig. \ref{subfig:scenario2}) to highly complex scenarios where the drones must communicate (e.g., Fig. \ref{subfig:scenario3}) to successfully avoid each other. The employed scenarios are the following:
\begin{figure}[t]
    \centering
    \includegraphics[width=8cm]{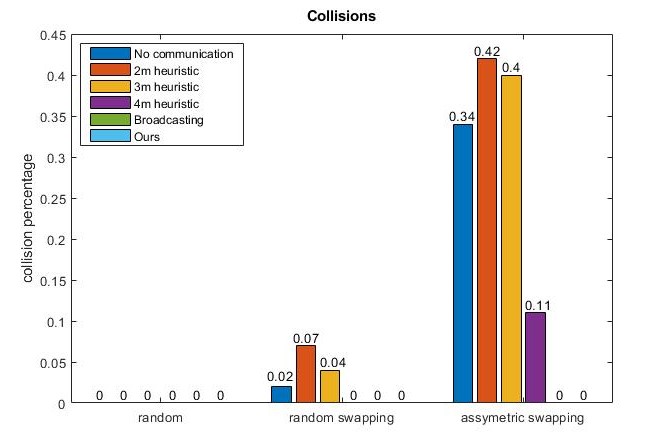}
    \caption{Collision avoidance performance for each scenario of the baseline methods vs. our learned-policy.}
    \label{fig:collisons}
\end{figure} 
\begin{itemize}
    \item \textbf{Random}: Each drone must to move to a random goal position. To ensure collision avoidance the drones must communicate their trajectory plans when crossing the path of another drone.
    \item \textbf{Random swapping}: each drone is requested to move to another drone's initial position.
    \item \textbf{Asymmetric swapping}: In this scenario we split the $\mathbb{R}^2$ x-y plane into four quadrants and initialize each drone in a different quadrant with random initial position. Then, each drone swaps positions with a drone from the diametrically opposed quadrant. If the drones do not communicate, collision is highly likely to occur.
\end{itemize}
Figure~\ref{fig:collisons} shows how the number of collisions varies per scenario considering a \textit{full-communication} and \textit{no-communication} policy.  
Depending on which scenario the agents are trained in a different communication policy may be learned. For instance, if an agent is trained only on the first scenario it will learn a \textit{no-communication} policy. In contrast, if only trained in the last it may learn to always communicate. Hence, we employ \textit{curriculum learning}~\cite{bengio2009curriculum}, training the agents first in a simple scenario, where no communication is needed, and continuously introducing more difficult and complex scenarios where the agents must learn \textit{when to communicate}.

\begin{figure}[t]
    \centering
    \includegraphics[width=8cm]{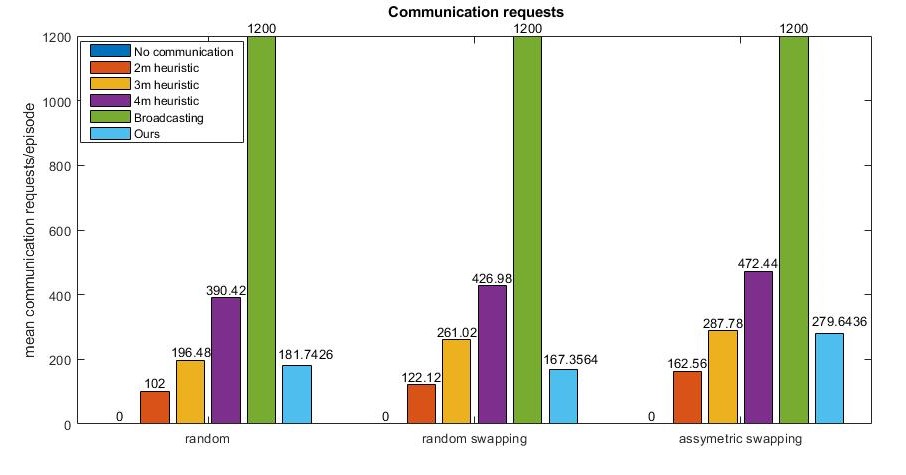}
    \caption{Number of communication requests for each scenario of the baseline methods vs. our learned-policy.}
    \label{fig:comm_req}
\end{figure}


\subsection{Performance Evaluation}
We compare our learned communication policy with two baseline approaches:
\begin{itemize}
    \item \emph{Full communication} (FC): At each time-step each drone broadcasts its trajectory plans.
    \item \emph{No communication} (NC): The drones never exchange their trajectory plans and a Constant Velocity model is used by each drone to infer the others trajectories.
    \item \emph{A distance-based} communication policy ($\epsilon$-DBCP): If the distance between two agents distance is smaller than a threshold $\epsilon$ (in meters) then the agents broadcast their trajectory information.
    
\end{itemize}
Fig. \ref{fig:comm_req} shows the number of collisions per scenario for each communication policy. In the first two scenarios the number of collisions is zero for any baseline. This demonstrates that for these two scenarios the simplified constant velocity model is enough and no-communication is required. In contrast, for the Random and Asymmetric scenarios the number of collisions raises significantly if the drones do not communicate. Yet, our learned policy achieved zero collisions in all scenarios. Moreover, Fig. \ref{fig:comm_req} demonstrates that our policy reduced the number of communications requests approximately 77\% while ensuring collision avoidance. In comparison with a 4-DBCP policy our method was able to reduce approximately 40\% the number of communications requests and the number of 11\% collisions to zero.
Finally, Fig. \ref{fig:scenarios} depicts the drone-swarm trajectories for each scenario. We can observe that our learned policy triggers communication either in the beginning of the motion or when the drones are in collision course.

\section{Conclusions}
In this paper, we have introduced an effective communication policy integrating the strengths of MARL and NMPC in collision avoidance tasks. Simulation results show that our policy learns "when" to request other agents to perform collision avoidance.
Furthermore, our method reduces the amount of communication requests significantly while ensuring collision-free motions. Future work will seek to scale our approach to a higher and variable number of agents and perform experimental results.

\bibliographystyle{IEEEtran}
\balance
\bibliography{iros_bib}
\end{document}